\documentclass[letterpaper]{article} 

\usepackage{spconf,amsmath,graphicx}
\usepackage{booktabs}
\usepackage{multirow}
\usepackage{amssymb}

\title{Real-time Semantic Scene Completion Via Feature Aggregation and Conditioned Prediction}
\name{Xiaokang Chen$^*$~\thanks{$^*$ Equal Contribution. 
	$\dagger$ This work is supported by the National Key Research and Development Program of China (2017YFB1002601, 2016QY02D0304), National Natural Science Foundation of China (61375022, 61403005, 61632003), Beijing Advanced Innovation Center for Intelligent Robots and Systems (2018IRS11), and PEK-SenseTime Joint Laboratory of Machine Vision.
 }, Yajie Xing$^*$, Gang Zeng}
\address{Peking University}
\begin{document}
%
\maketitle

\begin{abstract}
Semantic Scene Completion (SSC) aims to simultaneously predict the volumetric occupancy and semantic category of a 3D scene.
In this paper, we propose a real-time semantic scene completion method with a feature aggregation strategy and conditioned prediction module.
Feature aggregation fuses feature with different receptive fields and gathers context to improve scene completion performance.
And the conditioned prediction module adopts a two-step prediction scheme that takes volumetric occupancy as a condition to enhance semantic completion prediction.
We conduct experiments on three recognized benchmarks NYU, NYUCAD, and SUNCG.
Our method achieves competitive performance at a speed of 110 FPS on one GTX 1080 Ti GPU.
\end{abstract}
\section{Introduction}
\label{sec:intro}
We live in a 3D world, in which empty and occupied space is determined by the physical presence of objects. To understand the environment around us, we need to grasp the geometry and semantics of the scene simultaneously~\cite{chen2022context,chen2022d,chen2022group,tang2022not,chen2022conditional,meng2021conditional,chen2021semi,chen2020bi,chen20203d,chen2020real,tang2022compressible,tang2022point,JiaxiangTang2021JointII,MinZhong2023MaskGroupHP,QiangChen2022GroupDV,XinyuZhang2022CAEVC,JiaxiangTang2022RealtimeNR,JiaxiangTang2023DelicateTM,liu2022mpii,liu2021enhance,liu2023parallel}. Often, the single-view depth image obtained by depth sensor from a single perspective is incomplete, which means there are lots of areas occluded. To understand the 3D scene, we need to complete the invisible area according to the visible part.

Object shape completion has a long history in geometry processing. Some methods~\cite{nan2012search,shao2012interactive} complete the partial input of an object by matching it with 3D models from a large shape database. \cite{yuan2018pcn, stutz2018learning} propose a learning-based strategy that use a deep neural network with supervision to handle this task.

Although object shape completion could obtain a satisfactory result on a single object, it is hard to generalize to the completion of a whole scene. 
Therefore, Semantic Scene Completion (SSC) task is proposed by~\cite{song2017semantic-sscnet}. It aims to predict the volumetric occupancy and semantic category of a 3D scene simultaneously, which is proved to outperform either the completion or the semantic segmentation task. 

Considering the enormous computational and memory requirements of 3D CNN, ESSCNet~\cite{zhang2018efficient-esscnet} introduces Spatial Group Convolution (SGC) to divide input volume into different groups and conduct 3D sparse convolution on them. 
VVNet~\cite{guo2018view-vvnet} combines 2D CNN and 3D CNN with a differentiable projection layer to efficiently reduce the computational cost. 
Although these works are dedicated to reducing the computational cost of SSC, none of them has achieved real-time inference, leaving this task still far from practical application.
Dense prediction tasks like semantic segmentation or SSC usually need features combining both large-receptive-field semantic information and detailed local information to yield a better result.
In semantic segmentation, there are many works~\cite{lin2017refinenet, lee2017rdfnet, yu2018bisenet} that study feature aggregation strategies.
For semantic scene completion,
DDRNet~\cite{li2019rgbd-ddrnet} proposes a light-weight Dimensional Decomposition Residual (DDR) block and adopts a RDFNet\cite{lee2017rdfnet}-like feature fusion scheme to construct network.
It saves computation through DDR block and improves performance through feature aggregation, but still cannot become real-time.
Moreover, we notice that to do semantic-free completion is easier than to simultaneously finish completion and predict semantic categories, and the former can be served as a prior condition to the latter.
But no existing SSC methods have exploited this property.

\begin{figure*}[!ht]
  \centering
  \includegraphics[width=1.\linewidth]{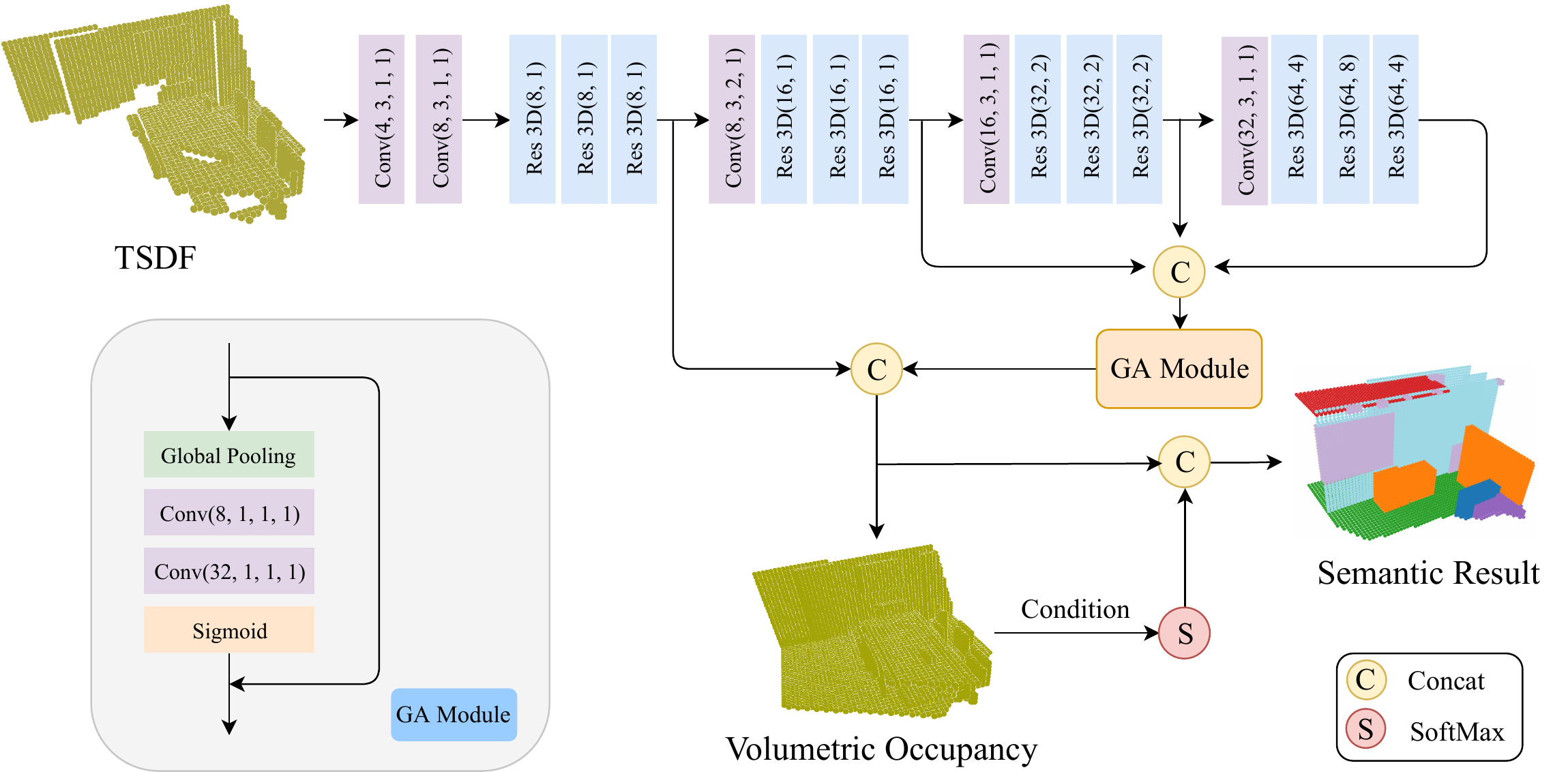}
  \caption{
  Overview of the proposed method. The convolution parameters are shown as (channel number, kernel size, stride, dilation). The Res 3D implies 3D Residual Block, and the parameters are shown as (channel number, dilation). 
  We propose an encoder with dilated convolution to obtain a large receptive field.
  Features extracted by the encoder are aggregated by a stage-wise aggregation strategy, where a Global Aggregation Module is also used to aggregate global context.
  Finally, the model adopts a step-wise conditioned prediction module to yield the final prediction.
  }
  \label{fig:arch}
  \vspace{-0.3cm}
\end{figure*}

In this paper, we propose a real-time semantic scene completion model, which runs at a speed of \textbf{110 FPS} with a highly competitive result.
We adopt a ResNet~\cite{he2016deep}-motivated network as our backbone.
To enlarge the receptive field, we utilize dilated convolution at the latter stages of the network.
We propose Global Aggregation Module to fuse global context feature and local feature, and a multi-level feature aggregation strategy containing the module to combine both global context and feature with different receptive fields.
Besides, we employ a two-step prediction scheme to exploit the condition relation between volumetric occupancy and semantic category. 
We first generate the binary occupancy prediction and yield semantic prediction according to both extracted feature and occupancy prediction.
Thus the occupancy could supply structural information for the semantic prediction, improving the performance. 
We conduct experiments on three public benchmarks (NYU Depth V2, NYUCAD, and SUNCG) and the experimental results validate the effectiveness of the proposed method.

\section{Proposed Method}
\label{sec:method}
Fig.~\ref{fig:arch} illustrates the overall structure of our model.
Given a TSDF (Truncated Signed Distance Function) data transformed from depth map, our model assign each voxel a semantic label $c\in\{c_0,c_1,\cdots,c_{N}\}$, where $N$ is the number of semantic categories and $c_0$ stands for empty voxels.
Our method consists of four key components: Dilated Convolution Encoder, Global Aggregation Module, Multi-level Feature Aggregation, and Conditioned Prediction.
In the following subsections, we describe these components in detail.

\subsection{Dilated Convolution Encoder}
As demonstrated in dense prediction tasks~\cite{yu2015dilatednet, chen2017deeplabv3, zhao2017pspnet}, a large receptive field is crucial to provide richer information for understanding the scene.
Considering 3D convolutions consume large memory and computation budget, we design a light-weight encoder that effectively enlarges the receptive field and extracts representative features.
We adopt a ResNet-like design to construct the encoder, in which two plain convolutions are used to process the raw input into low-level features, and cascaded residual blocks are used to refine the features.
The encoder downsamples the feature map and process high-dimensional features at a lower resolution to save computational cost.
We apply dilated convolutions at the latter downsampled stages to enlarge the receptive field and gather context information.
As shown in Fig.~\ref{fig:arch}, we adopt a ``multi-grid" dilation rate strategy similar to ~\cite{chen2017deeplabv3} in our encoder.
The effect of different dilation rates is studied in ablation studies.

\subsection{Global Aggregation Module}
The largest receptive field is the whole scene.
As proved in other semantic prediction tasks~\cite{chen2017deeplabv3,zhao2017pspnet}, global context that includes information in the whole scene is beneficial to enhancing local features and improving performance.
Therefore, we propose a Global Aggregation Module (GA Module) to aggregate global context into the local feature map.
As Fig.~\ref{fig:arch} shows, the global context is gathered through a 3D global pooling and incorporated into the local feature map with a channel-wise attention mechanism.

\subsection{Multi-level Feature Aggregation}
Although features with large receptive field and rich context have a strong capability for semantic prediction, they usually lack detailed high-frequency information, and thus degrades the dense prediction performance.
Therefore, we need to aggregate features with both large-range context and low-level details.
We propose a stage-wise aggregation strategy in our method.
We first concatenate feature maps generated by the three stages after downsampling.
Then a convolution is applied to reduce channel dimension.
After the high-level aggregated feature is obtained, we utilize the proposed Global Aggregation Module to moreover enhance the feature map.
Finally, we upsample the enhanced feature map back to the original resolution, concatenate it with a low-level feature map, and reduce the dimension through another convolution.

\subsection{Conditioned Prediction}

We propose that semantic scene completion can be divided into two steps: semantic-free scene completion and semantic-aware scene completion, where the former can serve as a prior condition to the latter.
Following this idea, we propose a conditioned prediction module to yield semantic scene completion results.
The model first predicts a semantic-free volumetric occupancy, which is formatted as a two-category dense labeling task.
The label $\hat{c}\in\{\hat{c}_0, \hat{c}_1\}$ of semantic-free completion is generated by
\begin{equation}
\left\{  
    \begin{gathered}
        \hat{c} = \hat{c}_0, \text{if } c=c_0 \\
        \hat{c} = \hat{c}_1, \text{otherwise}
    \end{gathered}
\right.  
\end{equation}
Then we concatenate the softmax-normalized prediction with the final feature map to introduce the prior condition and predict semantic scene completion results.
For training, the total loss is the summation of semantic-free completion loss and semantic completion loss:
\begin{equation}
    L(p,\hat{p},y,\hat{y}) = L_{s}(p, y) + L_{c}(\hat{p}, \hat{y})
\end{equation},
where $p,\hat{p}$ are respectively predictions for semantic completion and semantic-free completion, $y,\hat{y}$ are the corresponding ground truth, and $L_{s},  L_{c}$ are both softmax cross entropy loss.

\section{Experiments}
\label{sec:experiments}

\subsection{Datasets and Evaluation Metrics}
\noindent \textbf{Datasets.} We evaluate the proposed method on three benchmarks: NYU Depth V2~\cite{nyudv2} (which is denoted as NYU in the following),  NYUCAD~\cite{firman2016structured} and SUNCG~\cite{song2017semantic-sscnet}. \textbf{NYU} consists of 1449 indoor scenes that are captured via a Kinect sensor. There are 795 for training and 654 for test. We follow~\cite{song2017semantic-sscnet} and use the 3D annotated labels provided by~\cite{rock2015completing} for semantic scene completion task. \textbf{NYUCAD} uses the depth maps generated from the projections of the 3D annotations to address the misalignment of some label volumes and their corresponding depth maps. \textbf{SUNCG} is a synthetic dataset made by SSCNet~\cite{song2017semantic-sscnet}, which consists of 45622 indoor scenes. Follow \cite{song2017semantic-sscnet}, we adopt the same training/test split for our network training and evaluation. More specifically, 150K depth images and the corresponding ground-truth volumes for training and 470 pairs sampled from 170 non-overlap scenes for evaluation.

\noindent \textbf{Evaluation Metrics.}
We follow SSCNet~\cite{song2017semantic-sscnet} to use precision, recall and voxel-level intersection over union (IoU) as evaluation metrics. Specifically, two tasks are considered: semantic scene completion (SSC) and scene completion (SC). For the task of SSC, we evaluate the IoU of each object class on both observed and occluded voxels in the view frustum. For the task of SC, we treat all voxels as binary predictions and evaluate the binary IoU on occluded voxels in the view frustum.

\subsection{Implementation Details}
We use PyTorch framework to implement our experiments with a single GeForce GTX 1080 Ti GPU. We adopt mini-batch SGD with momentum to train our model with batch size $16$, momentum $0.9$ and weight decay $0.0005$. The initial learning rate is set to $0.1$ for NYU and NYUCAD, $0.02$ for SUNCG. We employ a poly learning rate policy where the initial learning rate is multiplied by $(1-\frac{iter}{max\_iter})^{0.9}$. We train our network for $300, 300$ and $20$ epochs for NYU, NYUCAD and SUNCG respectively.

\subsection{Ablation Studies}
\begin{table}[tbp]
\begin{center}
\caption{Ablation studies about different modules on the NYUCAD test set.}
\label{tab:ab-study}
\resizebox{0.99\columnwidth}{!}{
\begin{tabular}{cccccc}
\noalign{\smallskip}
\toprule
\textbf{Dilation} & \textbf{Feature Agg} & \textbf{GA} & \textbf{Condition} &  \textbf{SSC mIoU (\%)}  & \textbf{SC IoU (\%)} \\
\midrule
 & & & & 40.0 & 79.7 \\
\checkmark & & & & 42.7 & 80.8 \\
\checkmark & \checkmark & & & 43.0 & 81.3 \\
\checkmark & \checkmark & \checkmark & & 43.8 & 81.9 \\
\checkmark & \checkmark & \checkmark & \checkmark & \textbf{44.5} & \textbf{82.2} \\
\bottomrule
\end{tabular}
}
\vspace{-0.6cm}
\end{center}
\end{table}

We conduct ablation studies on NYUCAD to verify the proposed modules. From Table \ref{tab:ab-study}, we observe that each proposed module brings reasonable performance gain. Dilated convolution enlarges the receptive field of the network while keeping the size of feature maps unchanged, and boost the baseline by $2.7\%$ SSC mIoU. With the proposed multi-level aggregation strategy and the Global Aggregation Module, our network could jointly exploit feature maps with different receptive fields. This could improve the adaptability of our network to objects with different sizes. At last, the proposed occupancy condition supplies shape prior for the semantic prediction, thus boosting both the SSC mIoU and SC IoU.

\begin{table}[htp]
\begin{center}
\caption{Ablation studies about dilation rates on the NYUCAD test set.}
\label{tab:ab-study2}
\resizebox{0.8\columnwidth}{!}{
\begin{tabular}{ccc}
\noalign{\smallskip}
\toprule
\textbf{Dilation rate} & \textbf{SSC mIoU (\%)}  & \textbf{SC IoU (\%)} \\
\midrule
1,1,1,1,1,1 & 42.8 & 80.8 \\
2,2,2,4,4,4 & 43.2 & 81.6 \\
2,2,2,4,8,4 & \textbf{44.5} & \textbf{82.2} \\
\bottomrule
\end{tabular}
}
\vspace{-0.5cm}
\end{center}
\end{table}

To further verify the importance of dilated convolutions, we conduct ablation studies on different dilation rates. Results are listed in Table \ref{tab:ab-study2}. ``Dilation rate" implies $6$ dilation rates for the last $6$ 3D Residual Blocks in the encoder. From the table, we observe that with the increase of dilation rate, performance obtains a stable gain, which illustrates the importance of the large receptive field.

\subsection{Comparison with State-of-the-art Methods}
\noindent \textbf{Quantitative Results.} We compare the proposed method with other outstanding methods on three benchmarks: NYU, NYUCAD and SUNCG. Quantitative results are listed in Table \ref{tab:sota-nyu}, \ref{tab:sota-nyucad}, \ref{tab:sota-suncg} respectively. We observe that the proposed method obtains a consistently leading result on all three datasets, especially the SC IoU metric, which illustrates the effectiveness of the proposed modules.

\begin{table}[!htbp]
\begin{center}
\caption{Comparison with state-of-the-art methods on the NYU test set.}
\label{tab:sota-nyu}
\resizebox{0.8\columnwidth}{!}{
\begin{tabular}{ccc}
\noalign{\smallskip}
\toprule
\textbf{Method} & \textbf{SSC mIoU (\%)}  & \textbf{SC IoU (\%)} \\
\midrule
SSCNet~\cite{song2017semantic-sscnet} & 24.7 & 55.1 \\
VVNet~\cite{guo2018view-vvnet} & 32.9 & 61.1 \\
SATNet~\cite{liu2018see-satnet} & 34.4 & 60.6 \\
TS3D~\cite{garbade2018two-ts3d} & 30.4 & 60.4 \\
DDRNet~\cite{li2019rgbd-ddrnet} & 30.4 & 61.0 \\
\bottomrule
Ours & \textbf{34.4} & \textbf{73.4} \\
\bottomrule
\end{tabular}
}
\vspace{-0.5cm}
\end{center}
\end{table}

\begin{table}[ht]
\begin{center}
\caption{Comparison with state-of-the-art methods on the NYUCAD test set.}
\label{tab:sota-nyucad}
\resizebox{0.8\columnwidth}{!}{
\begin{tabular}{ccc}
\noalign{\smallskip}
\toprule
\textbf{Method} &  \textbf{SSC mIoU (\%)}  & \textbf{SC IoU (\%)} \\
\midrule
SSCNet~\cite{song2017semantic-sscnet} & 40.0 & 73.2 \\
VVNet~\cite{guo2018view-vvnet} &  \ & 80.3 \\
TS3D~\cite{garbade2018two-ts3d} &  42.1 & 74.2 \\
DDRNet~\cite{li2019rgbd-ddrnet} &  42.8 & 79.4 \\
\bottomrule
Ours & \textbf{44.5} & \textbf{82.2} \\
\bottomrule
\end{tabular}
}
\vspace{-0.5cm}
\end{center}
\end{table}

\begin{table}[ht]
\begin{center}
\caption{Comparison with state-of-the-art methods on the SUNCG test set.}
\label{tab:sota-suncg}
\resizebox{0.8\columnwidth}{!}{
\begin{tabular}{ccc}
\noalign{\smallskip}
\toprule
\textbf{Method} &  \textbf{SSC mIoU (\%)}  & \textbf{SC IoU (\%)} \\
\midrule
SSCNet~\cite{song2017semantic-sscnet} &  46.4 & 73.5 \\
VVNet~\cite{guo2018view-vvnet} & 66.7 & 84.0 \\
ESSCNet~\cite{zhang2018efficient-esscnet} & \textbf{70.5} & 84.5 \\
SATNet~\cite{liu2018see-satnet} & 64.3 & 78.5 \\
\bottomrule
Ours & 63.5 & \textbf{84.8} \\
\bottomrule
\end{tabular}
}
\vspace{-0.5cm}
\end{center}
\end{table}

\begin{table}[ht]
\begin{center}
\caption{The inference speed and GPU memory usage of the proposed method. All results are acquired on a GTX 1080 Ti GPU and evaluated on the NYU~\cite{nyudv2} test set.}
\label{tab:efficiency}
\resizebox{0.99\columnwidth}{!}{
\begin{tabular}{cccccc}
\noalign{\smallskip}
\toprule
\textbf{Method} & \textbf{Params (k)} & \textbf{FLOPs (G)}  & \textbf{Speed (FPS)}  & \textbf{Memory (M)} & \textbf{SSC-mIoU(\%)} \\
\midrule
SSCNet~\cite{song2017semantic-sscnet} & 930 & 163.8 & 0.7 & 5305 & 24.7\\
DDRNet~\cite{li2019rgbd-ddrnet} & 195 & 27.2 & 1.5 & 1829 & 30.4\\
Ours & \textbf{65} & \textbf{1.6} & \textbf{109.8} & \textbf{655} & \textbf{34.4} \\
\bottomrule
\end{tabular}
}
\vspace{-0.3cm}
\end{center}
\end{table}

\begin{figure}[!htb]
  \centering
  \includegraphics[width=1.\linewidth]{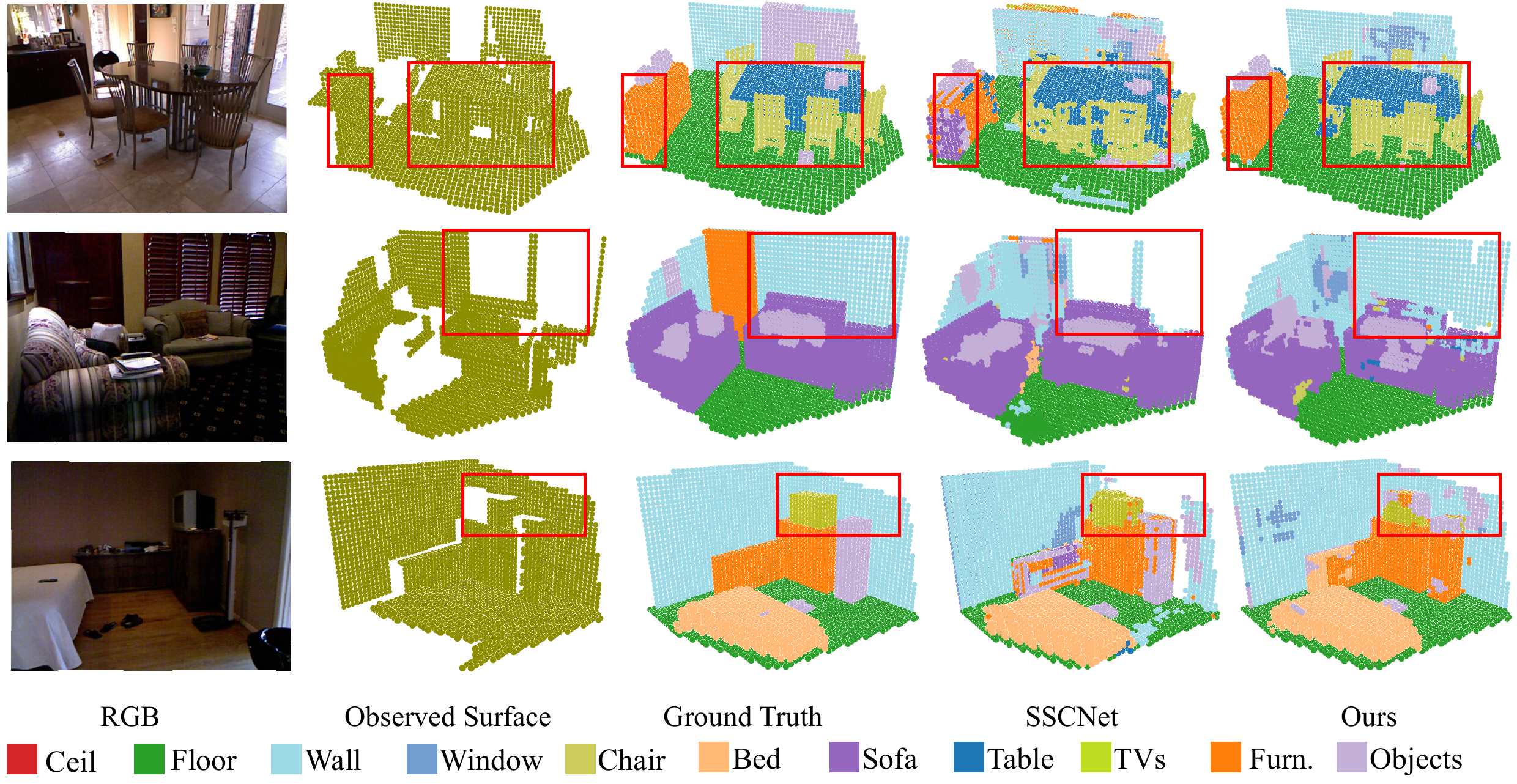}
  \caption{
  Comparison with SSCNet~\cite{song2017semantic-sscnet} on the NYUCAD test set. Best viewed in color.
  }
  \label{fig:sota}
  \vspace{-0.3cm}
\end{figure}

\noindent \textbf{Qualitative Results.}
Qualitative results are visualized in Figure \ref{fig:sota}. In the first row, we observe that the prediction of SSCNet lacks too many details (such as tables and chairs) because they didn't exploit low-level feature. In the second and the third rows, large areas of walls are missing in the predictions of SSCNet. As a comparison, the proposed method completes these areas well and precisely understands the semantics of them, because we combine features with different receptive fields, and the conditioned volumetric occupancy supplies the structure prior for the final prediction.

\noindent \textbf{Efficiency Analysis.} We analyze the computational efficiency of the proposed method following~\cite{li2019rgbd-ddrnet}. Statistics are listed in Table \ref{tab:efficiency}. As shown in the table, the proposed method has significantly fewer parameters and FLOPs than DDRNet~\cite{li2019rgbd-ddrnet}, while achieving much faster speed (about 73x faster) with a quite large performance gain.

\section{Conclusion}
\label{sec:conclusion}
In this paper, we propose a real-time semantic scene completion model with feature aggregation and conditioned prediction.
We propose a light-weight encoder with a large receptive field, and propose a Global Aggregation Module and a stage-wise feature aggregation strategy to fuse both high-level context-rich features and low-level detailed features.
And we propose a conditioned prediction module to predict volumetric occupancy and semantic completion step-by-step, exploiting volumetric occupancy as a prior condition for semantic completion prediction.
We conduct experiments on NYU, NYUCAD, and SUNCG to validate the effectiveness of our method.
Our method achieves a highly competitive performance at the speed of 110 FPS.

\small
\bibliographystyle{IEEEbib}
\bibliography{main}

\begin{thebibliography}{10}

\bibitem{chen2022context}
Xiaokang Chen, Mingyu Ding, Xiaodi Wang, Ying Xin, Shentong Mo, Yunhao Wang,
  Shumin Han, Ping Luo, Gang Zeng, and Jingdong Wang,
\newblock ``Context autoencoder for self-supervised representation learning,''
\newblock {\em arXiv preprint arXiv:2202.03026}, 2022.

\bibitem{chen2022d}
Xiaokang Chen, Jiahui Chen, Yan Liu, and Gang Zeng,
\newblock ``D$^3$etr: Decoder distillation for detection transformer,''
\newblock {\em arXiv preprint arXiv:2211.09768}, 2022.

\bibitem{chen2022group}
Qiang Chen, Xiaokang Chen, Jian Wang, Haocheng Feng, Junyu Han, Errui Ding,
  Gang Zeng, and Jingdong Wang,
\newblock ``Group detr: Fast detr training with group-wise one-to-many
  assignment,''
\newblock {\em arXiv preprint arXiv:2207.13085}, vol. 1, no. 2, 2022.

\bibitem{tang2022not}
Jiaxiang Tang, Xiaokang Chen, Jingbo Wang, and Gang Zeng,
\newblock ``Not all voxels are equal: Semantic scene completion from the
  point-voxel perspective,''
\newblock in {\em Proceedings of the AAAI Conference on Artificial
  Intelligence}, 2022, vol.~36, pp. 2352--2360.

\bibitem{chen2022conditional}
Xiaokang Chen, Fangyun Wei, Gang Zeng, and Jingdong Wang,
\newblock ``Conditional detr v2: Efficient detection transformer with box
  queries,''
\newblock {\em arXiv preprint arXiv:2207.08914}, 2022.

\bibitem{meng2021conditional}
Depu Meng, Xiaokang Chen, Zejia Fan, Gang Zeng, Houqiang Li, Yuhui Yuan, Lei
  Sun, and Jingdong Wang,
\newblock ``Conditional detr for fast training convergence,''
\newblock in {\em Proceedings of the IEEE/CVF International Conference on
  Computer Vision}, 2021, pp. 3651--3660.

\bibitem{chen2021semi}
Xiaokang Chen, Yuhui Yuan, Gang Zeng, and Jingdong Wang,
\newblock ``Semi-supervised semantic segmentation with cross pseudo
  supervision,''
\newblock in {\em Proceedings of the IEEE/CVF Conference on Computer Vision and
  Pattern Recognition}, 2021, pp. 2613--2622.

\bibitem{chen2020bi}
Xiaokang Chen, Kwan-Yee Lin, Jingbo Wang, Wayne Wu, Chen Qian, Hongsheng Li,
  and Gang Zeng,
\newblock ``Bi-directional cross-modality feature propagation with
  separation-and-aggregation gate for rgb-d semantic segmentation,''
\newblock in {\em Computer Vision--ECCV 2020: 16th European Conference,
  Glasgow, UK, August 23--28, 2020, Proceedings, Part XI}. Springer, 2020, pp.
  561--577.

\bibitem{chen20203d}
Xiaokang Chen, Kwan-Yee Lin, Chen Qian, Gang Zeng, and Hongsheng Li,
\newblock ``3d sketch-aware semantic scene completion via semi-supervised
  structure prior,''
\newblock in {\em Proceedings of the IEEE/CVF Conference on Computer Vision and
  Pattern Recognition}, 2020, pp. 4193--4202.

\bibitem{chen2020real}
Xiaokang Chen, Yajie Xing, and Gang Zeng,
\newblock ``Real-time semantic scene completion via feature aggregation and
  conditioned prediction,''
\newblock in {\em 2020 IEEE International Conference on Image Processing
  (ICIP)}. IEEE, 2020, pp. 2830--2834.

\bibitem{tang2022compressible}
Jiaxiang Tang, Xiaokang Chen, Jingbo Wang, and Gang Zeng,
\newblock ``Compressible-composable nerf via rank-residual decomposition,''
\newblock {\em arXiv preprint arXiv:2205.14870}, 2022.

\bibitem{tang2022point}
Jiaxiang Tang, Xiaokang Chen, Jingbo Wang, and Gang Zeng,
\newblock ``Point scene understanding via disentangled instance mesh
  reconstruction,''
\newblock in {\em Computer Vision--ECCV 2022: 17th European Conference, Tel
  Aviv, Israel, October 23--27, 2022, Proceedings, Part XXXII}. Springer, 2022,
  pp. 684--701.

\bibitem{JiaxiangTang2021JointII}
Jiaxiang Tang, Xiaokang Chen, and Gang Zeng,
\newblock ``Joint implicit image function for guided depth super-resolution,''
\newblock {\em arXiv: Computer Vision and Pattern Recognition}, 2021.

\bibitem{MinZhong2023MaskGroupHP}
Min Zhong, Xinghao Chen, Xiaokang Chen, Gang Zeng, and Yunhe Wang,
\newblock ``Maskgroup: Hierarchical point grouping and masking for 3d instance
  segmentation,''
\newblock 2023.

\bibitem{QiangChen2022GroupDV}
Qiang Chen, Jian Wang, Chuchu Han, Shan Zhang, Zexian Li, Xiaokang Chen, Jiahui
  Chen, Xiaodi Wang, Shuming Han, Gang Zhang, Haocheng Feng, Kun Yao, Junyu
  Han, Errui Ding, and Jingdong Wang,
\newblock ``Group detr v2: Strong object detector with encoder-decoder
  pretraining,''
\newblock 2022.

\bibitem{XinyuZhang2022CAEVC}
Xinyu Zhang, Jiahui Chen, Junkun Yuan, Qiang Chen, Jian Wang, Xiaodi Wang,
  Shumin Han, Xiaokang Chen, Jimin Pi, Kun Yao, Junyu Han, Errui Ding, and
  Jingdong Wang,
\newblock ``Cae v2: Context autoencoder with clip target,''
\newblock 2022.

\bibitem{JiaxiangTang2022RealtimeNR}
Jiaxiang Tang, Kaisiyuan Wang, Hang Zhou, Xiaokang Chen, Dongliang He, Tianshu
  Hu, Jingtuo Liu, Gang Zeng, and Jingdong Wang,
\newblock ``Real-time neural radiance talking portrait synthesis via
  audio-spatial decomposition,''
\newblock 2022.

\bibitem{JiaxiangTang2023DelicateTM}
Jiaxiang Tang, Hang Zhou, Xiaokang Chen, Tianshu Hu, Errui Ding, Jingdong Wang,
  and Gang Zeng,
\newblock ``Delicate textured mesh recovery from nerf via adaptive surface
  refinement,''
\newblock 2023.

\bibitem{liu2022mpii}
Yan Liu, Sanyuan Chen, Yazheng Yang, and Qi~Dai,
\newblock ``Mpii: Multi-level mutual promotion for inference and
  interpretation,''
\newblock in {\em Proceedings of the 60th Annual Meeting of the Association for
  Computational Linguistics (Volume 1: Long Papers)}, 2022, pp. 7074--7084.

\bibitem{liu2021enhance}
Yan Liu and Yazheng Yang,
\newblock ``Enhance long text understanding via distilled gist detector from
  abstractive summarization,''
\newblock {\em arXiv preprint arXiv:2110.04741}, 2021.

\bibitem{liu2023parallel}
Yan Liu, Xiaokang Chen, and Qi~Dai,
\newblock ``Parallel sentence-level explanation generation for real-world
  low-resource scenarios,''
\newblock {\em arXiv preprint arXiv:2302.10707}, 2023.

\bibitem{nan2012search}
Liangliang Nan, Ke~Xie, and Andrei Sharf,
\newblock ``A search-classify approach for cluttered indoor scene
  understanding,''
\newblock {\em TOG}, vol. 31, no. 6, pp. 137, 2012.

\bibitem{shao2012interactive}
Tianjia Shao, Weiwei Xu, Kun Zhou, Jingdong Wang, Dongping Li, and Baining Guo,
\newblock ``An interactive approach to semantic modeling of indoor scenes with
  an rgbd camera,''
\newblock {\em TOG}, vol. 31, no. 6, pp. 136, 2012.

\bibitem{yuan2018pcn}
Wentao Yuan, Tejas Khot, David Held, Christoph Mertz, and Martial Hebert,
\newblock ``Pcn: Point completion network,''
\newblock in {\em 3DV}. IEEE, 2018, pp. 728--737.

\bibitem{stutz2018learning}
David Stutz and Andreas Geiger,
\newblock ``Learning 3d shape completion from laser scan data with weak
  supervision,''
\newblock in {\em CVPR}, 2018, pp. 1955--1964.

\bibitem{song2017semantic-sscnet}
Shuran Song, Fisher Yu, Andy Zeng, Angel~X Chang, Manolis Savva, and Thomas
  Funkhouser,
\newblock ``Semantic scene completion from a single depth image,''
\newblock in {\em CVPR}, 2017, pp. 1746--1754.

\bibitem{zhang2018efficient-esscnet}
Jiahui Zhang, Hao Zhao, Anbang Yao, Yurong Chen, Li~Zhang, and Hongen Liao,
\newblock ``Efficient semantic scene completion network with spatial group
  convolution,''
\newblock in {\em ECCV}, 2018, pp. 733--749.

\bibitem{guo2018view-vvnet}
Yu-Xiao Guo and Xin Tong,
\newblock ``View-volume network for semantic scene completion from a single
  depth image,''
\newblock in {\em IJCAI}, 2018, pp.~--.

\bibitem{lin2017refinenet}
Guosheng Lin, Anton Milan, Chunhua Shen, and Ian~D. Reid,
\newblock ``Refinenet: Multi-path refinement networks for high-resolution
  semantic segmentation,''
\newblock in {\em {CVPR}}. 2017, pp. 5168--5177, {IEEE} Computer Society.

\bibitem{lee2017rdfnet}
Seungyong Lee, Seong{-}Jin Park, and Ki{-}Sang Hong,
\newblock ``Rdfnet: {RGB-D} multi-level residual feature fusion for indoor
  semantic segmentation,''
\newblock in {\em {ICCV}}. 2017, pp. 4990--4999, {IEEE} Computer Society.

\bibitem{yu2018bisenet}
Changqian Yu, Jingbo Wang, Chao Peng, Changxin Gao, Gang Yu, and Nong Sang,
\newblock ``Bisenet: Bilateral segmentation network for real-time semantic
  segmentation,''
\newblock in {\em {ECCV} {(13)}}. 2018, vol. 11217 of {\em Lecture Notes in
  Computer Science}, pp. 334--349, Springer.

\bibitem{li2019rgbd-ddrnet}
Jie Li, Yu~Liu, Dong Gong, Qinfeng Shi, Xia Yuan, Chunxia Zhao, and Ian Reid,
\newblock ``Rgbd based dimensional decomposition residual network for 3d
  semantic scene completion,''
\newblock in {\em CVPR}, 2019, pp.~--.

\bibitem{he2016deep}
Kaiming He, Xiangyu Zhang, Shaoqing Ren, and Jian Sun,
\newblock ``Deep residual learning for image recognition,''
\newblock in {\em CVPR}, 2016, pp. 770--778.

\bibitem{yu2015dilatednet}
Fisher Yu and Vladlen Koltun,
\newblock ``Multi-scale context aggregation by dilated convolutions,''
\newblock {\em CoRR}, vol. abs/1511.07122, 2015.

\bibitem{chen2017deeplabv3}
Liang{-}Chieh Chen, George Papandreou, Florian Schroff, and Hartwig Adam,
\newblock ``Rethinking atrous convolution for semantic image segmentation,''
\newblock {\em CoRR}, vol. abs/1706.05587, 2017.

\bibitem{zhao2017pspnet}
Hengshuang Zhao, Jianping Shi, Xiaojuan Qi, Xiaogang Wang, and Jiaya Jia,
\newblock ``Pyramid scene parsing network,''
\newblock in {\em {CVPR}}. 2017, pp. 6230--6239, {IEEE} Computer Society.

\bibitem{nyudv2}
Nathan Silberman, Derek Hoiem, Pushmeet Kohli, and Rob Fergus,
\newblock ``Indoor segmentation and support inference from rgbd images,''
\newblock in {\em ECCV}, 2012.

\bibitem{firman2016structured}
Michael Firman, Oisin Mac~Aodha, Simon Julier, and Gabriel~J Brostow,
\newblock ``Structured prediction of unobserved voxels from a single depth
  image,''
\newblock in {\em CVPR}, 2016, pp. 5431--5440.

\bibitem{rock2015completing}
Jason Rock, Tanmay Gupta, Justin Thorsen, JunYoung Gwak, Daeyun Shin, and Derek
  Hoiem,
\newblock ``Completing 3d object shape from one depth image,''
\newblock in {\em ICCV}, 2015, pp. 2484--2493.

\bibitem{liu2018see-satnet}
Shice Liu, Yu~Hu, Yiming Zeng, Qiankun Tang, Beibei Jin, Yinhe Han, and Xiaowei
  Li,
\newblock ``See and think: Disentangling semantic scene completion,''
\newblock in {\em NIPS}, 2018, pp. 261--272.

\bibitem{garbade2018two-ts3d}
Martin Garbade, Johann Sawatzky, Alexander Richard, and Juergen Gall,
\newblock ``Two stream 3d semantic scene completion,''
\newblock 2019.

\end{thebibliography}

\end{document}